%% file: main.tex
\documentclass{article}

\PassOptionsToPackage{numbers, compress}{natbib}
\usepackage[final]{neurips_2021}

\usepackage{algorithm,amsfonts,amsmath,amssymb,bm,booktabs,enumerate,graphicx,hyperref,microtype,nicefrac,url,wrapfig,xcolor}
\usepackage[noend]{algorithmic}
\usepackage[T1]{fontenc}
\usepackage[utf8]{inputenc}

\include{defs}

\title{Adaptive Risk Minimization:\\Learning to Adapt to Domain Shift}

\author{%
    Marvin Zhang$^{*1}$, Henrik Marklund$^{*2}$, Nikita Dhawan\thanks{equal contribution}~~$^1$,\\\textbf{Abhishek Gupta}$^1$, \textbf{Sergey Levine}$^1$, \textbf{Chelsea Finn}$^2$\\\\
    $^1$ UC Berkeley, $^2$ Stanford University\\
}

\begin{document}

\maketitle
\setcounter{footnote}{0}

\begin{abstract}

A fundamental assumption of most machine learning algorithms is that the training and test data are drawn from the same underlying distribution. However, this assumption is violated in almost all practical applications: machine learning systems are regularly tested under \emph{distribution shift}, due to changing temporal correlations, atypical end users, or other factors. In this work, we consider the problem setting of domain generalization, where the training data are structured into domains and there may be multiple test time shifts, corresponding to new domains or domain distributions. Most prior methods aim to learn a single robust model or invariant feature space that performs well on all domains. In contrast, we aim to learn models that \emph{adapt} at test time to domain shift using unlabeled test points. Our primary contribution is to introduce the framework of adaptive risk minimization~(ARM), in which models are directly optimized for effective adaptation to shift by learning to adapt on the training domains. Compared to prior methods for robustness, invariance, and adaptation, ARM methods provide performance gains of 1-4\% test accuracy on a number of image classification problems exhibiting domain shift.

\end{abstract}

\section{Introduction}
\label{sec:intro}

The standard assumption in empirical risk minimization~(ERM) is that the data distribution at test time will match the training distribution. When this assumption does not hold, i.e., when there is \emph{distribution shift}, the performance of standard ERM methods can deteriorate significantly~\citep{quinonero09nips,lazer14science}.

\begin{wrapfigure}{r}{0.4\linewidth}
    \vspace{-1em}
    \centering
    \includegraphics[width=\linewidth]{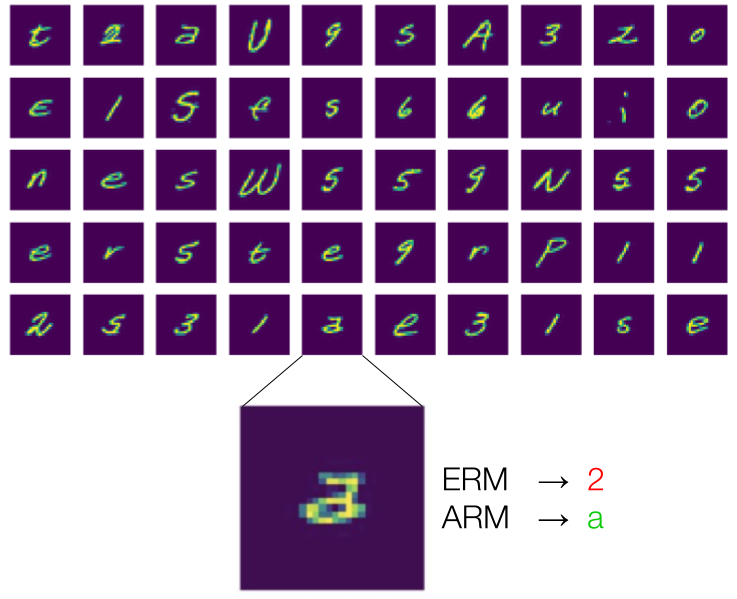}
    \caption{An example of ambiguous data points in handwriting classification, evaluated quantitatively in \autoref{sec:experiments}.}
    \label{fig:examples}
    \vspace{-2em}
\end{wrapfigure}

As an example which we study quantitatively in \autoref{sec:experiments}, consider a handwriting classification model that, after training on data from past users, is deployed to new end users. Each new user represents a new test distribution that differs from the training distribution. Thus, each test setting involves dealing with shift. In \autoref{fig:examples}, we visualize a batch of 50 examples from a test user, and we highlight an ambiguous example which may be either a ``2'' (written with a loop) or an ``a'' (in the double-storey style) depending on the user's handwriting. Due to the biases in the training data, an ERM trained model incorrectly classifies this example as ``2''. However, we can see that the batch of images from this test user contains other examples of ``2'' (written without loops) and ``a'' (also double-storey) from this user. Can we somehow leverage this unlabeled data to better handle test shifts caused by new users?

Any framework that aims to address this question must use additional assumptions beyond the ERM setting, and many such frameworks have been proposed~\citep{quinonero09nips}. One commonly used assumption within several frameworks, such as domain generalization~\citep{blanchard11nips,gulrajani21iclr}, is that the training data are provided in \emph{domains} and distributions at test time will represent new domains. The example above neatly fits this description if we equate users with domains -- we would be assuming that the training data are organized by users and that the model will be tested separately on new users, and these are reasonable assumptions. Constructing training domains in practice is generally accomplished by using meta-data, which exists for many commonly used datasets. Thus, this domain assumption is applicable for a wide range of realistic distribution shift problems (see, e.g., \citet{koh21icml}).

However, prior benchmarks for domain generalization and similar settings typically center around \emph{invariances} -- i.e., in these benchmarks, there is a consistent input-output relationship across all domains, and the goal is to learn this relationship while ignoring the spurious correlations within the domains (see, e.g., \citet{gulrajani21iclr}). Thus, prior methods aim for generalization to shifts by discovering this relationship, through techniques such as robust optimization and learning an invariant feature space~\citep{li18cvpr,arjovsky19,sagawa20iclr}. These methods are appealing in that they make minimal assumptions about the information provided at test time -- in particular, they do not require test labels, and the learned model can be immediately applied to predict on a single point. Nevertheless, these methods also have limitations, such as in dealing with problems where the input-output relationship varies across domains, e.g., the handwriting classification example above.

In this paper, we instead focus on methods that aim to \emph{adapt} at test time to domain shift. To do so, we study problems in which it is both feasible and helpful (and perhaps even necessary) to assume access to a batch or stream of inputs at test time. Leveraging this test assumption does not require labels for any test data and is feasible in many practical setups. For example, for handwriting classification, we do not access only single handwritten characters from an end user, but rather collections of characters such as sentences or paragraphs. Unlabeled adaptation has been shown empirically to be useful for distribution shift problems~\citep{sun20icml,schneider20neurips,wang21iclr}, such as for dealing with image corruptions~\citep{hendrycks19aiclr}. Taking inspiration from these findings, we propose and evaluate on a number of problems, detailed in \autoref{sec:experiments}, for which adaptation is beneficial in dealing with domain shift.

Our main contribution is to introduce the framework of adaptive risk minimization~(ARM), which proposes the following objective: optimize the model such that it can maximally leverage the unlabeled adaptation phase to handle domain shift. To do so, we instantiate a set of methods that, given a set of training domains, \emph{meta-learns} a model that is adaptable to these domains. These methods are straightforward extensions of existing meta-learning approaches, thereby demonstrating that tools from the meta-learning toolkit can be readily adapted to tackle domain shift. Our experiments in \autoref{sec:experiments} test on several image classification problems, derived from benchmarks for federated learning~\citep{caldas19flneurips} and image classifier robustness~\citep{hendrycks19aiclr}, in which training and test domains share structure that can be leveraged for improved performance. These testbeds are also a contribution of our work, as we believe these problems can supplement existing benchmarks which, almost exclusively, are designed with invariance in mind~\citep{arjovsky19,peng19iccv,gulrajani21iclr}. We also evaluate on the \textsc{Wilds} suite of distribution shift problems~\citep{koh21icml}, which have been curated to faithfully represent important real world problems. Empirically, we demonstrate that the proposed ARM methods, by leveraging meta-training and test time adaptation, are often able to outperform prior state-of-the-art methods by \mbox{1-4\%} test accuracy.

\section{Related Work}
\label{sec:related}

A number of prior works have studied distribution shift in various forms~\citep{quinonero09nips}. In this section, we review prior work in domain generalization, group robustness, meta-learning, and adaptation.

\textbf{Invariance and robustness to domains.} As discussed above, a number of frameworks leverage training domains to address test time shift. The terminology in prior work is scattered and, depending on the application, includes terms such as ``groups'', ``datasets'', ``subpopulations'', and ``users''; in this work, we adopt the term ``domains'' which we believe is an appropriate unifying term. A number of testbeds for this problem setting have been proposed for image classification, including generalizing to new datasets~\citep{fang13iccv}, new image types~\citep{li17iccv,peng19iccv}, and underrepresented demographics~\citep{sagawa20iclr}.

Prior benchmarks typically assume the existence of a consistent input-output relationship across domains that is learnable by the specified model, thus motivating methods such as learning an invariant feature space~\citep{li18cvpr,li18eccv,arjovsky19} or optimizing for worst case group performance~\citep{hu18icml,sagawa20iclr}. In particular, methods for domain generalization -- sometimes referred to as multi-source domain adaptation~\citep{sun15if} or zero shot domain adaptation~\citep{yang15iclr} -- have largely focused on learning invariant features~\citep{ganin15icml,sun16aaai,li18cvpr,li18eccv,peng19iccv}. \citet{gulrajani21iclr} provide a comprehensive survey of domain generalization benchmarks and find that, surprisingly, ERM is competitive with the state of the art across all the benchmarks considered. In \autoref{sec:supp-bench}, we discuss this finding as well as the performance of an ARM method on this benchmark suite. In \autoref{sec:experiments}, we identify different problems for which adaptation is helpful, and we find that, on these problems, ARM methods consistently outperform ERM and other non adaptive methods for robustness and invariance.

\textbf{Meta-learning.} Meta-learning~\citep{schmidhuber87,bengio92optimality,thrun98l2l,hochreiter01icann} has been most extensively studied in the context of few shot \emph{labeled} adaptation~\citep{santoro16icml,vinyals16nips,ravi17iclr,finn17icml,snell17nips}. Our aim is not to address few-shot recognition problems, nor to propose a novel meta-learning algorithm, but rather to extend meta-learning paradigms to problems requiring unlabeled adaptation, with the primary goal of tackling distribution shift. This aim differs from previous work in meta-learning for domain generalization~\citep{li18aaaai,dou19neurips}, which seek to meta-train models for non adaptive generalization performance. We discuss in \autoref{sec:arm} how paradigms such as contextual meta-learning~\citep{garnelo18icml,requeima19neurips} can be readily extended using the ARM framework.

Some other meta-learning methods adapt using both labeled and unlabeled data, either in the semi supervised learning setting~\citep{ren18iclr,zhang18neurips,li19neurips} or the transductive learning setting~\citep{nichol18,liu19iclr,antoniou19neurips,hu20iclr}. These works all assume access to labeled data for adaptation, whereas we propose methods and problems for purely unlabeled adaptation. Prior works in meta-learning for unlabeled adaptation include \citet{yu18rss}, who adapt a policy to imitate human demonstrations in the context of robotic learning; \citet{metz19iclr}, who meta-learn an update rule for unsupervised representation learning, though they still require labels to learn a predictive model; and \citet{alet21neurips}, who meta-learn adaptive models based on task specific unsupervised objectives. Unlike these prior works, we propose a general framework for tackling distribution shift problems by meta-learning unsupervised adaptation strategies. This framework simplifies the extension of meta-learning paradigms to these problems, encapsulates previous approaches such as the gradient based meta-learning approach of \citet{yu18rss}, and sheds light on how to improve existing strategies such as adaptation via batch normalization~\citep{li17iclrw}.

\textbf{Adaptation to shift.} Unlabeled adaptation has primarily been studied separately from meta-learning. Domain adaptation is a prominent framework that assumes access to test examples at \emph{training} time~\citep{csurka17,wilson20tist}, similar to transductive learning~\citep{vapnik98slt}. As such, most domain adaptation methods consider the problem setting where there is a single test distribution~\citep{shimodaira00jspi,daume07acl,gong12cvpr,ganin15icml,tzeng17cvpr,carlucci17iccv}, and some of these methods are difficult to apply to problems where there are multiple test distributions. Certain domain adaptation methods have also been applied in the domain generalization setting, such as methods for learning invariant features~\citep{ganin15icml,sun16aaai,li18cvpr}, and we compare to these methods in \autoref{sec:experiments}.

Adaptive methods for domain generalization include \citet{muandet13icml} and \citet{kumagai18}, who propose a method similar to one of the ARM methods described below. We compare to a version of this method in \autoref{sec:supp-exp}. \citet{blanchard11nips} and \citet{blanchard21jmlr} provide a theoretic study of domain generalization and establish favorable generalization bounds for models that can adapt to domain shift at test time. We summarize some of these results in \autoref{sec:prelim}. In comparison, our work establishes a framework that makes explicit the connection between adaptation to domain shift and meta-learning, allowing us to devise new methods in a straightforward and principled manner. These methods are amenable to expressive models such as deep neural networks, which enables us to propose and evaluate on problems with raw image observations.

Test time adaptation has also been studied for dealing with label shift~\citep{royer15cvpr,lipton18icml,sulc19iccv} and crafting favorable inductive biases for the domain of interest. For image classification, techniques such as normalizing via the test inputs~\citep{li17iclrw} and optimizing self-supervised surrogate losses~\citep{sun20icml} have proven effective for adapting to image corruptions~\citep{hendrycks19aiclr}. We compare to these prior methods in \autoref{sec:experiments} and empirically demonstrate the advantage of using training domains to learn how to adapt.

\section{Preliminaries and Notation}
\label{sec:prelim}

In this section, we discuss the domain generalization problem setting and formally describe adaptive models. In \autoref{sec:arm}, we discuss how adaptive models can be meta-trained via the ARM objective and approach, and we instantiate ARM methods which we empirically evaluate in \autoref{sec:experiments}.

Let $\x\in\inspace$ and $y\in\outspace$ represent the input and output, respectively. We can formalize the domain generalization problem setting using the following data generation process~\citep{blanchard21jmlr}: first, a joint data distribution $p_{\xy}$ is sampled from a set of distributions $\P_{\xy}$, and then some data points are sampled from $p_{\xy}$.\footnote{Formally, the number of points sampled is another random variable with support over the positive integers.} We refer to each $p_{\xy}$ as a domain, e.g., a particular dataset or user, thus $\P_{\xy}$ represents the set of all possible domains. We assume that the training dataset is composed of data from $S$ runs of this generative process, organized by domain. An equivalent characterization which we will use for clarity is that, within the training set, there are $S$ domains, and each data point $(\x^{(i)},y^{(i)})$ is annotated with a domain label $z^{(i)}$. Each $z^{(i)}$ is an integer that takes on a value between $1$ and $S$, indicating which $p_{\xy}$ generated the $i$-th training point (though, of course, we do not have access to, or knowledge of, $p_{\xy}$ itself). At test time, there may be multiple evaluation settings, where each setting is considered separately and contains only \emph{unlabeled} data sampled via a new run of the same generative process. This data may represent, e.g., a new dataset or user, and the test domains are likely to be distinct from the training domains when $|\P_{\xy}|$ is large or infinite.

Our formal goal is to optimize for expected performance, e.g., classification accuracy, at test time. To do so, let us first consider predictive models of the form \mbox{$\fullmodel:\inspace\times\P_{\x}\to\outspace$}, where the model $\fullmodel$ takes in not just an input $\x$ but also the marginal input distribution $p_{\x}\in\P_{\x}$ that $\x$ was sampled from. We refer to $\fullmodel$ as an \emph{adaptive} model, as it has the opportunity to use $p_{\x}$ to adapt its predictions on $\x$. The underlying assumption is that $p_{\x}$ provides information about $p_{y\vert\x}$, i.e., $p_{\x}$ is used as a surrogate input in place of $p_{\xy}$. In the worst case, if $p_{\x}$ and $p_{y\vert\x}$ are sampled independently, then the model does not benefit at all from knowing $p_{\x}$. In many problems, however, we expect knowledge about $p_{\x}$ to be useful, e.g., for resolving ambiguity as in the handwriting classification example in \autoref{sec:intro}.

Theoretically, when $p_{\x}$ provides information about $p_{y\vert\x}$, and when training and test domains are drawn from the same distribution over $\P_{\xy}$, we can establish favorable generalization bounds for the expected performance of $\fullmodel$ in adapting to domain shift at test time. We can formalize this as follows. First, define a \emph{prediction model} to be a non adaptive model of the form \mbox{$\model:\inspace\to\outspace$}, and define the risk for a prediction model $\model$ and loss function $\ell$, under a data distribution $p_{\xy}$, as
\[
\mathcal{R}(\model,p_{\xy})\triangleq\E_{p_{\xy}}\left[\ell(\model(\x),y)\right]\,.
\]
Further, define the Bayes optimal risk for $\ell$ under $p_{\xy}$ as
\[
\mathcal{R}^\star(p_{\xy})\triangleq\min_\model\mathcal{R}(\model,p_{\xy})\,.
\]
Let $\mu$ denote the distribution on $\P_{\xy}$ from which training and test domains $p_{\xy}$ are sampled. To avoid overlapping terms, define the \emph{adaptive risk} for an adaptive model $\fullmodel$ and $\ell$, under $\mu$, to be
\begin{equation}
\mathcal{E}(\fullmodel,\mu)\triangleq\E_\mu\left[\E_{p_{\xy}}\left[\ell(\fullmodel(\x,p_{\x}),y)\right]\right]\,.\label{eq:ar}
\end{equation}
We state the following result from \citet{blanchard21jmlr}, which details a condition on $\mu$ under which $\mathcal{E}$ is a strongly principled objective for learning adaptive models.

\paragraph{Lemma 9 from \citet{blanchard21jmlr}.} Let $\fullmodel^\star$ denote a minimizer of $\mathcal{E}$ for the given $\mu$. If $\mu$ is a distribution on $\P_{\xy}$ such that $\mu$-almost surely it holds that $p_{y\vert\x}=M(p_{\x})$ for some deterministic mapping $M$, then for $\mu$-almost all $p_{\xy}$, we have
\[
\mathcal{R}(\fullmodel^\star(\cdot,p_{\x}),p_{\xy})=\mathcal{R}^\star(p_{\xy})\implies\mathcal{E}(\fullmodel^\star,\mu)=\E_\mu\left[\mathcal{R}^\star(p_{\xy})\right]\,.
\]
In other words, an adaptive model which minimizes the adaptive risk $\mathcal{E}$ coincides with a Bayes optimal decision function for $p_{\xy}$, \emph{for $\mu$-almost all domains $p_{\xy}$}.

\paragraph{Remark.} The required condition on $\mu$ -- that $p_{y\vert\x}$ is determined by $p_{\x}$ -- holds if, and only if, an expert (or oracle) is able to correctly label inputs from a given domain provided only information about the input distribution. This condition holds for the testbeds proposed in this paper, those in \citet{gulrajani21iclr}, and those in \textsc{Wilds}~\citep{koh21icml}. The condition does not hold for, e.g., standard few shot learning testbeds, where it is possible for two domains with identical input distributions to shuffle their label orderings differently~\citep{vinyals16nips}. Thus, these problems are outside the scope of this work.

This result provides strong justification for learning adaptive models $\fullmodel$ by minimizing the adaptive risk $\mathcal{E}$. However, a practical instantiation of this approach requires some approximations. First, we do not know and cannot input $p_{\x}$ to $\fullmodel$ in most cases. Instead, we instantiate $\fullmodel$ such that it takes in a batch of inputs $\x_1,\ldots,\x_K$, all from the same domain, where $K$ can vary. $\fullmodel$ makes predictions on the whole batch, which also serves as an empirical approximation (i.e., a histogram) $\hat{p}_{\x}$ of $p_{\x}$~\citep{blanchard21jmlr}. In our exposition, we will assume that a batch of unlabeled points is available at test time for adaptation. However, we also experiment in \autoref{sec:experiments} with the \emph{streaming} setting where the test inputs are observed one at a time and adaptation occurs incrementally.

Notice that, if we instead passed in an approximation $\hat{p}_{\xy}$ of $p_{\xy}$ to the model, such as a batch of \emph{labeled} data $(\x_1,y_1),\ldots,(\x_K,y_K)$, then this setup would resemble the standard few shot meta-learning problem \citep{vinyals16nips}. Formally, a meta-learning model takes in both an input $\x$ and $\hat{p}_{\xy}$, which approximates the distribution that $\x$ was sampled from and thus can be used to adapt the prediction on $\x$. Compared to our problem setting, the meta-learning formalism can tackle a wider range of problems but also requires more restrictive assumptions, specifically, labels at test time via $\hat{p}_{\xy}$. \emph{Transductive meta-learning} methods further assume that, in addition to $\hat{p}_{\xy}$, a full batch of inputs $\x_1,\ldots,\x_K$ is passed into the model, which allows for better estimation of the input distribution $p_{\x}$~\citep{nichol18,liu19iclr,hu20iclr}. The model then makes predictions on this entire batch. In meta-learning terminology, $\hat{p}_{\xy}$ and $\x_1,\ldots,\x_K$ are often referred to as the \emph{support} and \emph{query}, respectively. Therefore, another interpretation of the adaptive models that we study in this work is that they resemble transductive meta-learning models, but they are given only the unlabeled query and not the labeled support set.

In the next section, we expand on this connection to develop the ARM framework, which then allows us to bring forward tools from meta-learning to tackle domain shift problems.

\section{Adaptive Risk Minimization}
\label{sec:arm}

In this section, we formally describe the ARM framework, which defines an objective for training adaptive models to tackle domain shift. Furthermore, we propose a general meta-learning algorithm as well as specific methods for optimizing the ARM objective. In \autoref{sec:experiments}, we test these ARM methods on problems for which unlabeled adaptation can be leveraged for better test performance.

\subsection{Devising the ARM objective}
\label{sec:objective}

We wish to learn an adaptive model \mbox{$\fullmodel:\inspace^K\to\outspace^K$} to tackle domain shift. As noted, meta-learning methods for labeled adaptation study a similar form of model, and a common approach in many of these methods is to define $\fullmodel$ such that it is composed of two parts: first, a \emph{learner} which ingests the data and produces parameters, and second, a \emph{prediction model} which uses these parameters to make predictions~\citep{vinyals16nips,finn17icml}. We will follow a similar strategy which, as we will discuss in \autoref{sec:algorithm}, allows us to easily extend and design meta-learning methods towards our goal.

In particular, we will decompose the model $\fullmodel$ into two modules: a standard prediction model $\model({\cdot\,};\params):~\inspace\to\outspace$, that is parameterized by $\params\in\paramsspace$ and predicts $y$ given $\x$, and an \emph{adaptation model} $\learner({\cdot\,,\cdot\,};\learnerparams):~\paramsspace\times\inspace^K\to\paramsspace$, which is parameterized by $\learnerparams$. $\learner$ takes in the prediction model parameters $\params$ and $K$ unlabeled data points and uses the $K$ points to produce adapted parameters $\params'$. This is analogous to the learner in meta-learning, however, $\learner$ adapts the model parameters using only unlabeled data. We defer the discussion of how to instantiate $\learner$ to \autoref{sec:algorithm}.\footnote{For some meta-learning methods, the learner does not take as input the unadapted model parameters~\citep{vinyals16nips}, and we also devise some methods of this form. In the formalism above, these methods simply ignore the input $\params$.}

The ARM objective is to optimize $\learnerparams$ and $\params$ such that $\learner$ can adapt $\model$ using unlabeled data sampled according to a particular domain $z$. This can be expressed as the optimization problem
\begin{gather}
    \min_{\params,\learnerparams}\hat{\mathcal{E}}(\params,\learnerparams)=\E_{p_z}\left[\E_{p_{\xy\vert z}}\left[\frac{1}{K}\sum_{k=1}^K\ell(\model(\x_k;\params'),y_k)\right]\right]\,,\text{ where }~\params'=\learner(\params,\x_1,\ldots,\x_K;\learnerparams)\,.\label{eq:arm}
\end{gather}

Note that $\hat{\mathcal{E}}$ is the empirical form of the adaptive risk in \autoref{eq:ar} for the form of $\fullmodel$ we have defined. Mimicking the generative process from \autoref{sec:prelim} that we assume generated the training data, $p_z$ is a categorical distribution over $\{1,\ldots,S\}$ which places uniform probability mass on each training domain, and $p_{\xy\vert z}$ assigns uniform probability to only the training points within a particular domain. As we have established theoretically, we expect the trained models to perform well at test time if the test domains are sampled independently and identically -- i.e., from the same distribution over $\P_{\xy}$ -- as the training domains. In practice, similar to how meta-learned few shot classification models are evaluated on new and unseen meta-test classes~\citep{vinyals16nips,finn17icml}, we empirically show in \autoref{sec:experiments} that the trained models can generalize to test domains that are not sampled identically to the training domains.

\subsection{Optimizing the ARM objective}
\label{sec:algorithm}

\begin{wrapfigure}{r}{0.53\linewidth}
\vspace{-2.5em}
\begin{minipage}{\linewidth}
\begin{algorithm}[H]
\small
\linespread{1.2}\selectfont
\caption{Meta-Learning for ARM}
\label{alg:arm}
\vspace{.25em}
{\tt // Training procedure}
\vspace{-.15em}
\begin{algorithmic}[1]
    \REQUIRE{\# training steps $T$, batch size $K$, learning rate $\eta$}
    \STATE \textbf{Initialize:} $\params,\learnerparams$
    \FOR{$t=1,\ldots,T$}
        \STATE Sample $z$ uniformly from training domains
        \STATE Sample $(\x_k,y_k)\sim p({\cdot\,,\cdot\,}|z)$ for $k=1,\ldots,K$
        \STATE $\params'\leftarrow\learner(\params,\x_1,\ldots,\x_K;\learnerparams)$
        \STATE $(\params,\learnerparams)\leftarrow(\params,\learnerparams)-\eta\nabla_{(\params,\learnerparams)}\sum_{k=1}^K\ell(\model(\x_k;\params'),y_k)$
    \ENDFOR
\end{algorithmic}
\vspace{.5em}
{\tt // Test time adaptation procedure}
\begin{algorithmic}[1]
    \setcounter{ALC@line}{\numexpr6}
    \REQUIRE{$\params$, $\learnerparams$, test batch $\x_1,\ldots,\x_K$}
    \STATE $\params'\leftarrow\learner(\params,\x_1,\ldots,\x_K;\learnerparams)$
    \STATE $\hat{y}_k\leftarrow\model(\x_k;\params')$ for $k=1,\ldots,K$
\end{algorithmic}
\end{algorithm}
\end{minipage}
\vspace{-1.25em}
\end{wrapfigure}

\autoref{alg:arm} presents a general meta-learning approach for optimizing the ARM objective. As described above, $\learner$ outputs updated parameters $\params'$ using an unlabeled batch of data (line~5). This mimics the adaptation procedure at test time, where we do not assume access to labels (lines~7-8). However, the training update itself does rely on the labels (line~6). We assume that $\learner$ is differentiable with respect to its input $\params$ and $\learnerparams$, thus we use gradient updates on both $\params$ and $\learnerparams$ to optimize for \emph{post adaptation} performance on a mini batch of data sampled according to a particular domain $z$. In practice, we also sample mini batches of domains, rather than just one domain (as written in line~3), to provide a better gradient signal for optimizing $\learnerparams$ and $\params$.

Together, \autoref{eq:arm} and \autoref{alg:arm} shed light on a number of ways to devise methods for solving the ARM problem. First, we can extend meta-learning paradigms to the ARM problem setting, and any paradigm in which the adaptation model $\learner$ can be augmented to operate on unlabeled data is readily applicable. As an example, we propose the ARM-CML method, which is inspired by recent works in contextual meta-learning~(CML)~\citep{garnelo18icml,requeima19neurips}. Second, we can enhance prior unlabeled adaptation methods by incorporating a meta-training phase that allows the model to better leverage the adaptation. To this end, we propose the ARM-BN method, based on the general approach of adapting using batch normalization~(BN) statistics of the test inputs~\citep{li17iclrw,schneider20neurips,kaku20,nado20}. Third, we can incorporate existing methods for meta-learning unlabeled adaptation to solve domain shift problems. We demonstrate this by proposing the ARM-LL method, which is based on the robotic imitation learning method from \citet{yu18rss} which adapts via a learned loss~(LL). All of these methods are straightforward extensions of existing meta-learning and adaptation methods, and this is intentional -- we aim to show how existing tools can be readily adapted to tackle domain generalization problems. We summarize the methods here and refer the reader to \autoref{sec:supp-app} for complete details.

\textbf{ARM-CML.} In ARM-CML, the parameters $\learnerparams$ of $\learner$ define the weights of a \emph{context network} \mbox{$\contextnet({\cdot\,};\learnerparams):\inspace\to\mathbb{R}^D$}, parameterized by the adaptation model parameters $\learnerparams$. We also instantiate the model with a \emph{prediction network} \mbox{$\prednet({\cdot\,,\cdot\,};\params):\inspace\times\mathbb{R}^D\to\outspace$}, parameterized by $\params$. When given a mini batch of inputs, $\contextnet$ processes each example $\x_k$ in the mini batch separately and outputs $\context_k\in\mathbb{R}^D$ for $k=1,\ldots,K$, which are averaged together into a \emph{context} $\context=\frac{1}{K}\sum_{k=1}^K\context_k$. $D$ is a hyperparameter, and in our experiments, we choose $D$ to be the dimensionality of $\x$, such that we can concatenate each image $\x_k$ and the context $\context$ along the channel dimension to produce the input to $\prednet$. In other words, $\prednet$ processes each $\x_k$ separately to produce an estimate of the output $\hat{y}_k$, but it additionally receives $\context$ as input. In this way, $\contextnet$ can provide information about the entire batch of $K$ unlabeled data points to $\prednet$ for predicting the correct outputs.

Note that the difference between ARM-CML and prior contextual meta-learning approaches is that, in prior approaches, the context network processes both inputs and outputs to produce each $\context_k$. ARM-CML is designed for the domain generalization setting in which we do not assume access to labels at test time, thus we meta-train for unlabeled adaptation performance at training time.

\textbf{ARM-BN.} ARM-BN is a particularly simple method that is applicable for any model $\model$ that has BN layers~\citep{ioffe15icml}. Practically, training $\model$ via ARM-BN follows the same protocol as \citet{ioffe15icml} except for two key differences: first, the training batches are sampled from a single domain, rather than from the entire dataset, and second, the normalization statistics are recomputed at test time rather than using a training running average. As noted, this second difference has been explored by several works as a method for test time adaptation, but the first difference is novel to ARM-BN. Following \autoref{alg:arm}, ARM-BN defines a meta-training procedure in which $\model$ learns to adapt -- i.e., compute normalization statistics -- using batches of training points sampled from the same domain. We empirically show in \autoref{sec:experiments} that, for problems where BN adaptation already has a favorable inductive bias, such as for image classification, further introducing meta-training boosts its performance. We believe that other test time adaptation methods, such as those based on optimizing surrogate losses~\citep{sun20icml,wang21iclr}, may similarly benefit from their corresponding meta-training procedures.

\begin{wrapfigure}{r}{0.5\linewidth}
    \vspace{-1.25em}
    \centering
    \includegraphics[width=\linewidth]{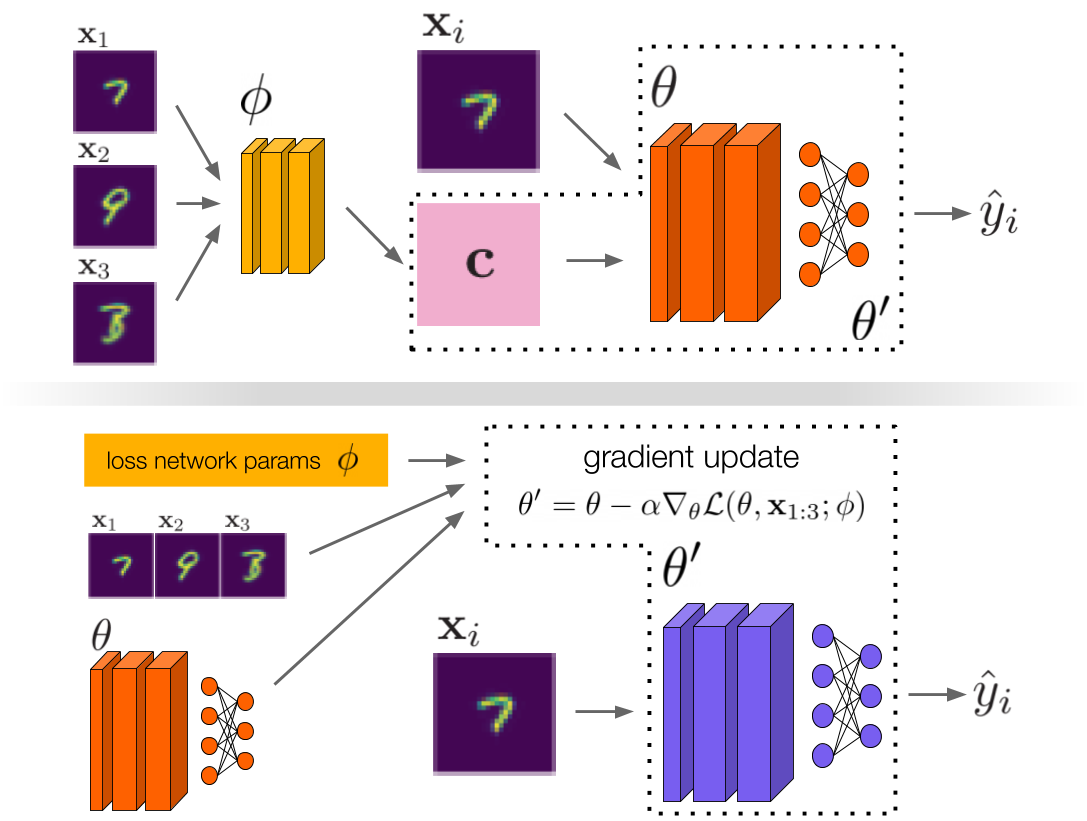}
    \caption{In the contextual approach~(top), $\x_1,\ldots,\x_K$ are summarized into a context $\context$, and we propose two methods for this summarization, either through a separate context network or using batch normalization activations in the model itself. $\context$ can then be used by the model to infer additional information about the input distribution. In the gradient based approach~(bottom), an unlabeled loss function $\mathcal{L}$ is used for gradient updates to the model parameters, in order to produce parameters that are specialized to the test inputs and can produce more accurate predictions.}
    \label{fig:methods}
    \vspace{-1.75em}
\end{wrapfigure}

At a high level, ARM-BN operates in a similar fashion to ARM-CML, thus we group these methods together into the umbrella of contextual approaches, shown in \autoref{fig:methods}~(top).  The interpretation of ARM-BN through the contextual approach is that $\learner$ replaces the running statistics used by standard BN with statistics computed on the batch of inputs, which then serves as the context $\context$. Thus, for ARM-BN, there is no context network, and $\learner$ has no parameters beyond the model parameters $\params$ involved in computing BN statistics. The model $\model$ is again specified via a prediction network $\prednet$, which must have BN layers. BN typically tracks a running average of the first and second moments of the activations in these layers, which are then used at test time. ARM-BN defines $\learner$ such that it swaps out these moments for the moments computed via the activations on the test batch, thus giving us adapted parameters $\params'$ if we view the moments as part of the model parameters. This method is remarkably simple, and in deep learning libraries such as PyTorch~\citep{paszke19neurips}, implementing ARM-BN involves changing a single line of code. However, as shown in \autoref{sec:experiments}, this method also performs very well empirically, and the adaptation effectiveness is further boosted by meta-training.

\textbf{ARM-LL.} ARM-LL, depicted in \autoref{fig:methods}~(bottom), follows the gradient based meta-learning paradigm~\citep{finn17icml} and learns parameters $\params$ that are amenable to gradient updates on a loss function in order to quickly adapt to a new problem. In other words, $\learner$ produces \mbox{$\params'=\params-\alpha\nabla_\params\mathcal{L}(\params,\x_1,\ldots,\x_K;\learnerparams)$}, where $\alpha$ is a hyperparameter. Note that the loss function $\mathcal{L}$ used in the gradient updates is different from the original supervised loss function $\ell$, in that it operates on only the inputs $\x$, rather than the input output pairs that $\ell$ receives. We follow the general implementation of this approach proposed in \citet{yu18rss}. We define $\model$ to produce output features $\mathbf{o}\in\mathbb{R}^{|\outspace|}$ that are used as logits when making predictions. We then define the unlabeled loss function $\mathcal{L}$ to be the composition of $\model$ and a \emph{loss network} $\lossnet({\cdot\,};\learnerparams):\mathbb{R}^{|\outspace|}\to\mathbb{R}$, which takes in the output features from $\model$ and produces a scalar. We use the $\ell_2$-norm of these scalars across the batch of inputs as the loss for updating $\params$. In other words,
\begin{gather*}
    \learner(\params,\x_1,\ldots,\x_K;\learnerparams)=\params-\alpha\nabla_\params\|\mathbf{v}\|_2\,,\text{ where }
    \mathbf{v}=[\lossnet(\model(\x_1;\params);\learnerparams),\ldots,\lossnet(\model(\x_K;\params);\learnerparams)]\,.
\end{gather*}

\section{Experiments}
\label{sec:experiments}

Our experiments are designed to answer the following questions:
\vspace{-.5em}
\begin{enumerate}[1.]
    \itemsep0em
    \item Do ARM methods learn models that can leverage unlabeled adaptation to tackle domain shift?
    \item How do ARM methods compare to prior methods for robustness, invariance, and adaptation?
    \item Can models trained via ARM methods adapt successfully in the streaming test setting?
\end{enumerate}

\subsection{Evaluation domains and protocol}

We propose four image classification problems, which we present below and describe in full detail in \autoref{sec:supp-setup}. We also present results on datasets from the \textsc{Wilds} benchmark~\citep{koh21icml} in \autoref{sec:wilds}. We believe that the problems we propose in this paper can supplement existing benchmarks for domain shift, which, as discussed above, are designed to test invariances. A key characteristic of the problems presented here is the potential for adaptation to improve test performance, and this differs from prior benchmarks such as the problems compiled by DomainBed~\citep{gulrajani21iclr}. In \autoref{sec:supp-bench}, we compare our testbeds to DomainBed and group robustness benchmarks, and we briefly discuss the results in \citet{gulrajani21iclr}, which also evaluate ARM-CML.

\textbf{Rotated MNIST.} We study a modified version of MNIST where images are rotated in 10 degree increments, from 0 to 130 degrees. We treat each rotation as a separate domain, i.e., a different value of $z$. We use only 108 training data points for each of the 2 smallest domains (120 and 130 degrees), and 324 points each for rotations 90 to 110, whereas the overall training set contains 32292 points. In this setting, we hypothesize that adaptation can specialize the model to specific domains, in particular the \emph{rare domains} in the training set. For each test evaluation, we generate images from the MNIST test set with a certain rotation. We measure both worst case and average accuracy across domains.

\textbf{Federated Extended MNIST (FEMNIST).} The extended MNIST~(EMNIST) dataset consists of images of handwritten uppercase and lowercase letters, in addition to digits~\citep{cohen17}. FEMNIST is the same dataset, but it also provides the meta-data of which user generated each data point~\citep{caldas19flneurips}. We treat each user as a domain. We measure each method's worst case and average accuracy across 35 test users, which are held out and thus \emph{disjoint} from the training users. As discussed in \autoref{sec:intro}, adaptation may help for this problem for specializing the model and resolving ambiguous data points.

\textbf{Corrupted image datasets.} CIFAR-10-C and Tiny ImageNet-C~\citep{hendrycks19aiclr} augment the CIFAR-10~\citep{krizhevsky09} and Tiny ImageNet test sets with common image corruptions that vary in type and severity. The original goal of these augmented test sets was to benchmark how well methods could handle these corruptions without access to \emph{any} corruptions during training~\citep{hendrycks19aiclr}. Thus, successful methods for these problems typically have relied on domain knowledge and heuristics designed specifically for image classification. For example, prior work has shown that carefully designed test time adaptation procedures are effective for these problems~\citep{sun20icml,schneider20neurips,wang21iclr}. One possible reason for this phenomenon is that convolutional networks are biased toward texture~\citep{geirhos19iclr}, which is distorted by corruptions, thus adaptation can help the model recover its performance for each corruption type.

We study whether meta-training for adaptation performance can improve upon these results. To do so, we modify the protocol from \citet{hendrycks19aiclr} to fit into the ARM problem setting by applying a set of 56 corruptions to the training data, and we define each corruption to be a domain. We use a \emph{disjoint} set of 22 corruptions for the test data, which are mostly of different types from the training corruptions (thus, not sampled identically), and we measure worst case and average accuracy across the test corruptions. This modification allows us to study, for both ARM and prior methods, whether seeing corruptions at training time can help the model deal with new corruptions at test time.

\subsection{Comparisons and ablations}

We compare the ARM methods against several prior methods designed for robustness, invariance, and adaptation. We describe the comparisons here and provide additional details in \autoref{sec:supp-setup}.

\textbf{Test time adaptation.} We evaluate the general approach of using test batches to compute BN statistics~\citep{li17iclrw,schneider20neurips,kaku20,nado20}, which we term BN adaptation. We also compare to test time training~(TTT)~\citep{sun20icml}, which adapts the model at test time using a self-supervised rotation prediction loss. These methods have previously achieved strong results for image classification, likely because they constitute favorable inductive biases for improving on the true classification task~\citep{sun20icml}.

\textbf{Ablations.} We also include ablations of the ARM-CML and ARM-LL methods, which sample training batches of unlabeled examples uniformly from the entire training set, rather than sampling from a single domain.\footnote{Note that the corresponding ablation of ARM-BN is simply the BN adaptation method.} These ``context ablation'' and ``learned loss ablation'' are similar to test time adaptation methods in that they do not require training domains, thus they allow us to evaluate whether or not meta-training on domain shifts is important for improved performance.

\textbf{Group robustness and invariance.} \citet{sagawa20iclr} recently proposed a state-of-the-art method for group robustness, and we refer to this approach as distributionally robust neural networks~(DRNN). Their work also evaluates a strong upweighting~(UW) baseline that samples uniformly from each group, and so we also evaluate this approach in our experiments. Additionally, we compare to domain adversarial neural networks~(DANN)~\citep{ganin15icml} and maximum mean discrepancy~(MMD) feature learning~\citep{li18cvpr}, two state-of-the-art methods for adversarial learning of invariant predictive features. For the \textsc{Wilds} datasets, we include the numbers reported in \citet{koh21icml} for DRNN and two other invariance methods, correlation alignment~(CORAL)~\citep{sun16aaai} and invariant risk minimization~(IRM)~\citep{arjovsky19}.

Robustness and invariance methods assume access to training domains but not test batches, whereas adaptation methods assume the opposite. Thus, at a high level, we can view the comparisons to these methods as evaluating the importance of each of these assumptions for the specified problems.

\def\*#1{\mathbf{#1}}
\def\+#1{\underline{\mathbf{#1}}}
\newcommand{\tc}[1]{\multicolumn{2}{c}{#1}}
\begin{table}[t]
    \centering
    \caption{Worst case~(WC) and average~(Avg) top 1 accuracy on all testbeds, where means and standard errors are reported across three separate runs of each method. Horizontal lines separate methods that make use of (from top to bottom): neither, training domains, test batches, or both. ARM methods consistently achieve greater robustness, measured by WC, and Avg performance compared to prior methods. $^*$UW is identical to ERM for CIFAR-10-C and Tiny ImageNet-C.\\\\
    These results have been updated from an earlier version of the paper, primarily for CIFAR-10-C, due to significant refactoring of the code, additional hyperparameter tuning for both the ARM methods and the prior methods, and efforts to standardize results across the authors' different computing environments and library versions. These results are reproducible from the publicly available code: \url{https://github.com/henrikmarklund/arm}.}
    \label{tab:results}
    \resizebox{\textwidth}{!}{%
    \begin{tabular}{lccccccccc}
        \toprule
                        & \tc{\textbf{MNIST}}                 & \tc{\textbf{FEMNIST}}               & \tc{\textbf{CIFAR-10-C}}            & \tc{\textbf{Tiny ImageNet-C}} \\
                          \cmidrule(lr){2-3}                    \cmidrule(lr){4-5}                    \cmidrule(lr){6-7}                    \cmidrule(lr){8-9}
        \textbf{Method} & \textbf{WC}      & \textbf{Avg}     & \textbf{WC}      & \textbf{Avg}     & \textbf{WC}      & \textbf{Avg}     & \textbf{WC}      & \textbf{Avg} \\
        \midrule
        ERM             & $74.5\pm1.4$     & $93.6\pm0.4$     & $62.4\pm0.4$     & $79.1\pm0.3$     & $54.1\pm0.3$     & $70.4\pm0.1$     & $20.3\pm0.5$     & $41.9\pm0.1$ \\
        \midrule
        UW$^*$          & $\*{80.3\pm1.2}$ & $\*{95.1\pm0.1}$ & $\*{65.7\pm0.7}$ & $80.3\pm0.6$     & ---              & ---              & ---              & --- \\
        DRNN            & $79.9\pm0.7$     & $\*{94.9\pm0.1}$ & $57.5\pm1.7$     & $76.5\pm1.2$     & $49.3\pm0.9$     & $65.7\pm0.5$     & $14.2\pm0.2$     & $31.6\pm1.0$ \\
        DANN            & $78.8\pm0.8$     & $\*{94.9\pm0.1}$ & $\*{65.4\pm1.0}$ & $\*{81.7\pm0.3}$ & $\*{53.9\pm2.2}$ & $\*{69.8\pm0.3}$ & $\*{20.4\pm0.7}$ & $\*{40.9\pm0.2}$ \\
        MMD             & $\*{82.4\pm0.9}$ & $\*{95.3\pm0.3}$ & $62.4\pm0.7$     & $79.8\pm0.4$     & $\*{52.2\pm0.3}$ & $\*{69.5\pm0.1}$ & $\*{19.7\pm0.2}$ & $40.1\pm0.1$ \\
        \midrule
        BN adaptation   & $78.0\pm0.3$     & $94.4\pm0.1$     & $65.7\pm1.5$     & $80.0\pm0.5$     & $60.6\pm0.3$     & $70.9\pm0.1$     & $26.5\pm0.3$     & $\*{42.8\pm0.0}$ \\
        TTT             & $\*{81.1\pm0.3}$ & $\*{95.4\pm0.1}$ & $\*{68.6\pm0.4}$ & $\*{84.2\pm0.1}$ & $\*{61.5\pm0.3}$ & $\*{71.7\pm0.5}$ & $\*{27.6\pm0.5}$ & $37.7\pm0.3$ \\
        CML ablation    & $63.5\pm1.8$     & $90.1\pm0.2$     & $61.8\pm0.8$     & $81.6\pm0.5$     & $58.8\pm0.1$     & $69.6\pm0.2$     & $26.3\pm0.6$     & $42.5\pm0.1$ \\
        LL ablation     & $\*{79.9\pm1.1}$ & $\*{95.0\pm0.3}$ & $64.1\pm1.6$     & $80.8\pm0.2$     & $\*{60.9\pm0.4}$ & $\*{71.3\pm0.0}$ & $21.6\pm2.1$     & $32.6\pm3.2$ \\
        \midrule
        ARM-CML         & $\*{88.0\pm0.8}$ & $\*{96.3\pm0.4}$ & $\+{70.9\pm1.4}$ & $\+{86.4\pm0.3}$ & $61.2\pm0.4$     & $70.3\pm0.2$     & $\+{29.1\pm0.4}$ & $\+{43.3\pm0.1}$ \\
        ARM-BN          & $83.3\pm0.5$     & $95.6\pm0.1$     & $64.5\pm3.2$     & $83.2\pm0.5$     & $\+{61.7\pm0.3}$ & $\*{72.4\pm0.3}$ & $\*{28.3\pm0.3}$ & $\+{43.3\pm0.1}$ \\
        ARM-LL          & $\+{88.9\pm0.8}$ & $\+{96.9\pm0.2}$ & $67.0\pm0.9$     & $84.3\pm0.7$     & $\*{61.2\pm0.7}$ & $\+{72.5\pm0.4}$ & $25.4\pm0.1$     & $35.7\pm0.4$ \\
        \bottomrule
    \end{tabular}
    }
    \vspace{0.25em}
\end{table}

\subsection{Quantitative evaluation and comparisons}
\label{sec:quantitative}

The results for the four proposed benchmarks are presented in \autoref{tab:results}. The best results, stratified by classes of methods, are bolded, with the single best result across all methods underlined. Across all of these problems, ARM methods increase both worst case and average accuracy compared to all other methods. ARM-CML performs well across all tasks, and despite its simplicity, ARM-BN achieves the best performance overall on the corrupted image testbeds, demonstrating the effectiveness of meta-training on top of an already strong adaptation procedure. BN adaptation and TTT are the strongest prior methods, as these adaptation procedures constitute inductive biases that are generally well suited for image classification. However, ARM methods are comparatively less reliant on favorable inductive biases and consistently attain better results. In general, we observe poor performance from robustness methods, varying performance from invariance methods, strong performance from adaptation methods, and the strongest performance from ARM methods.

\begin{figure}
    \centering
    \includegraphics[width=0.36\textwidth]{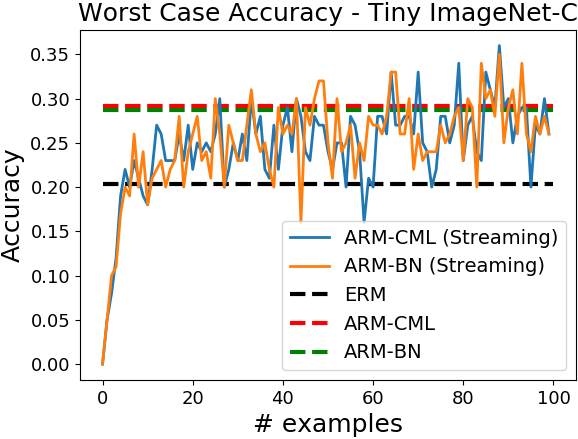}
    \hspace{2em}
    \includegraphics[width=0.3545\textwidth]{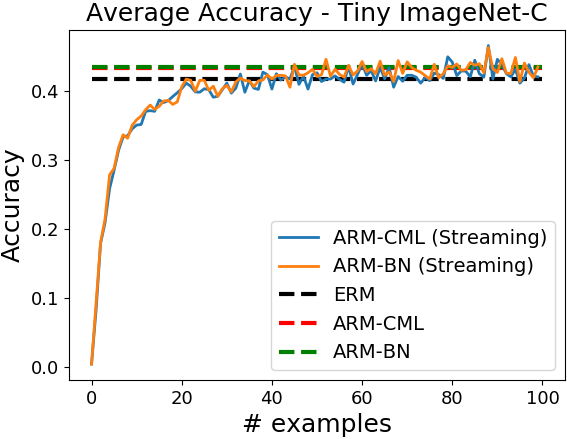}
    \caption{In the streaming setting, ARM methods reach strong performance on Tiny ImageNet-C after fewer than 50 data points, despite using training batch sizes of 100. This highlights that the trained models are able to adapt successfully in the standard streaming evaluation setting.}
    \label{fig:streaming}
\end{figure}

When we cannot access a batch of test points all at once, and instead the points are observed in a streaming fashion, we can augment the proposed ARM methods to perform sequential model updates. For example, ARM-CML and ARM-BN can update their average context and normalization statistics, respectively, after observing each new test point. In \autoref{fig:streaming}, we study this test setting for the Tiny ImageNet-C problem. We see that both models trained with ARM-CML and ARM-BN are able to achieve near their original worst case and average accuracy within observing 50 data points, well before the training batch size of 100. This result demonstrates that ARM methods are applicable for problems where test points must be observed one at a time, provided that the model is permitted to adapt using each point. We describe in detail how each ARM method can be applied to the streaming setting in \autoref{sec:supp-app}, and we provide streaming results on rotated MNIST in \autoref{sec:supp-exp}.

\subsection{\textsc{Wilds} results}
\label{sec:wilds}

\begin{table}[t]
    \centering
    \caption{Results on the \textsc{Wilds} image testbeds. Different methods are best suited for different problems, motivating the need for a wide range of methods. ARM-BN struggles on FMoW but performs well on the other datasets, in particular RxRx1.}
    \label{tab:wilds}
    \resizebox{\textwidth}{!}{%
    \begin{tabular}{lcccccccc}
        \toprule
                        & \tc{\textbf{iWildCam}}               & \textbf{Camelyon17} & \textbf{RxRx1}    & \tc{\textbf{FMoW}}                    & \tc{\textbf{PovertyMap}} \\
                          \cmidrule(lr){2-3}                     \cmidrule(lr){4-4}    \cmidrule(lr){5-5}  \cmidrule(lr){6-7}                     \cmidrule(lr){8-9}
        \textbf{Method} & \textbf{Acc}     & \textbf{Macro F1} & \textbf{Acc}        & \textbf{Acc}      & \textbf{WC Acc}   & \textbf{Avg Acc}  & \textbf{WC Pearson r} & \textbf{Pearson r} \\
        \midrule
        ERM             & $\*{71.6\pm2.5}$ & $31.0\pm1.3$      & $70.3\pm6.4$        & $29.9\pm0.4$      & $\*{32.3\pm1.25}$ & $\*{53.0\pm0.55}$ & $0.45\pm0.06$         & $0.78\pm0.04$ \\
        DRNN            & $\*{72.7\pm2.0}$ & $23.9\pm2.1$      & $68.4\pm7.3$        & $23.0\pm0.3$      & $\*{30.8\pm0.81}$ & $\*{52.1\pm0.5}$  & $0.39\pm0.06$         & $0.75\pm0.07$ \\
        CORAL           & $\*{73.3\pm4.3}$ & $\*{32.8\pm0.1}$  & $59.5\pm7.7$        & $28.4\pm0.3$      & $\*{31.7\pm1.24}$ & $50.5\pm0.36$     & $0.44\pm0.06$         & $0.78\pm0.05$ \\
        IRM             & $59.8\pm3.7$     & $15.1\pm4.9$      & $64.2\pm8.1$        & $8.2\pm1.1$       & $\*{30.0\pm1.37}$ & $50.8\pm0.13$     & $0.43\pm0.07$         & $0.77\pm0.05$ \\
        BN adaptation   & $46.4\pm1.0$     & $13.8\pm0.3$      & $\*{88.6\pm1.4}$    & $20.0\pm0.2$      & $30.2\pm0.26$     & $51.6\pm0.16$     & $0.39\pm0.17$         & $0.82\pm0.06$ \\
        ARM-BN          & $\*{70.3\pm2.4}$ & $23.2\pm2.7$      & $\*{87.2\pm0.9}$    & $\*{31.2\pm0.1}$  & $24.6\pm0.04$     & $42.0\pm0.21$     & $0.49\pm0.21$         & $0.84\pm0.05$ \\
        \bottomrule
    \end{tabular}
    }
    \vspace{0.25em}
\end{table}

Finally, we present results on the \textsc{Wilds} benchmark~\citep{koh21icml} in \autoref{tab:wilds}. We evaluate BN adaptation and ARM-BN on these testbeds. We see that, on these real world distribution shift problems, different methods perform well for different problems. CORAL, a method for invariance~\citep{sun16aaai}, performs best on the iWildCam animal classification problem~\citep{beery20}, whereas no methods outperform ERM by a significant margin on the FMoW~\citep{christie18cvpr} or PovertyMap~\citep{yeh20naturecomm} satellite imagery problems. ARM-BN performs particularly poorly on the FMoW problem. However, it performs well on PovertyMap and significantly improves performance on the RxRx1~\citep{taylor19iclr} problem of treatment classification from medical images. On the other medical imagery problem of Camelyon17~\citep{bandi18mi} tumor identification, adaptation in general boosts performance dramatically. These results indicate the need to consider a wide range of tools, including meta-learning and adaptation, for combating distribution shift.

\section{Discussion and Future Work}
\label{sec:disc}

We presented adaptive risk minimization~(ARM), a framework and problem formulation for learning models that can adapt in the face of domain shift at test time using only a batch of unlabeled test examples. We devised an algorithm and instantiated a set of methods for optimizing the ARM objective that meta-learns models that are adaptable to different domains of training data. Empirically, we observed that ARM methods consistently improve performance in terms of both average and worst case metrics, as compared to a number of prior approaches for handling domain shift.

Though we provided contextual meta-learning as a concrete example, a number of other meta-learning paradigms would also be interesting to extend to the ARM setting. For example, few shot generative modeling objectives would be a natural fit for unlabeled adaptation~\citep{edwards17iclr,hewitt18uai,wu19aaai}. Another exciting direction for future work is to explore the problem setting where domains are not provided at training time. As discussed in \autoref{sec:supp-exp}, in this setting, we can instead construct domains via unsupervised learning techniques. Similar to \citet{hsu19iclr}, one promising approach is to generate a diverse set of domains in order to learn generally effective adaptation strategies. Robustness and invariance methods cannot be used easily with multiple different groupings, learned or otherwise, as techniques such as group weighted loss functions~\citep{sagawa20iclr} and domain classifiers~\citep{ganin15icml} are not immediately extendable to this setup. Thus, ARM methods may be uniquely suited to be paired with domain learning.

\begin{ack}
MZ thanks Matt Johnson and Sharad Vikram for helpful discussions and was supported by an NDSEG fellowship. HM is funded by a scholarship from the Dr. Tech. Marcus Wallenberg Foundation for Education in International Industrial Entrepreneurship. AG was supported by an NSF graduate research fellowship. CF is a CIFAR Fellow in the Learning in Machines and Brains program. This research was supported by the DARPA Assured Autonomy and Learning with Less Labels programs.
\end{ack}

\bibliography{marvin}
\bibliographystyle{plainnat}

\clearpage
\appendix

\section{Broader Impacts}
\label{sec:supp-broader}

Though machine learning systems have been deployed in many real world domains with great success, data that is anomalous or structurally different from the training data still sometimes renders these systems unreliable, harmful, or even dangerous. It is necessary, in order to realize the full potential of machine learning ``in the wild'', to have effective methods for detecting, robustifying against, and adapting to distribution shift. The potential upsides of developing such methods are clear. Imagine systems for image classification that fix incorrect or offensive outputs by adapting to each end user, or self driving cars that can smoothly adapt to driving in a new setting. We believe our work is a small step toward the goal of adapting in the face of distribution shift.

However, there are also complications and downsides that must be considered. For example, it is important to understand the failure modes and theoretical limits to handling distribution shift, otherwise we may place ``false confidence'' in our deployed systems, which may be catastrophic. Our work does not address this aspect of the problem, though this is an important direction for future work. Perhaps more insidiously, this line of research may grant even greater capabilities to parties that are able to collect larger and larger datasets. Deep learning systems are capable of effectively learning from ever growing data, and as the training data grows, the system can be trained to better adapt to a wider range of potential shifts. Thus, it is imperative to continue to push for high quality open source datasets, so that we may democratize the tools of machine learning.

\section{More Details on the ARM Methods}
\label{sec:supp-app}

\begin{wrapfigure}{r}{0.58\linewidth}
    \vspace{2em}
    \centering
    \includegraphics[width=0.95\linewidth]{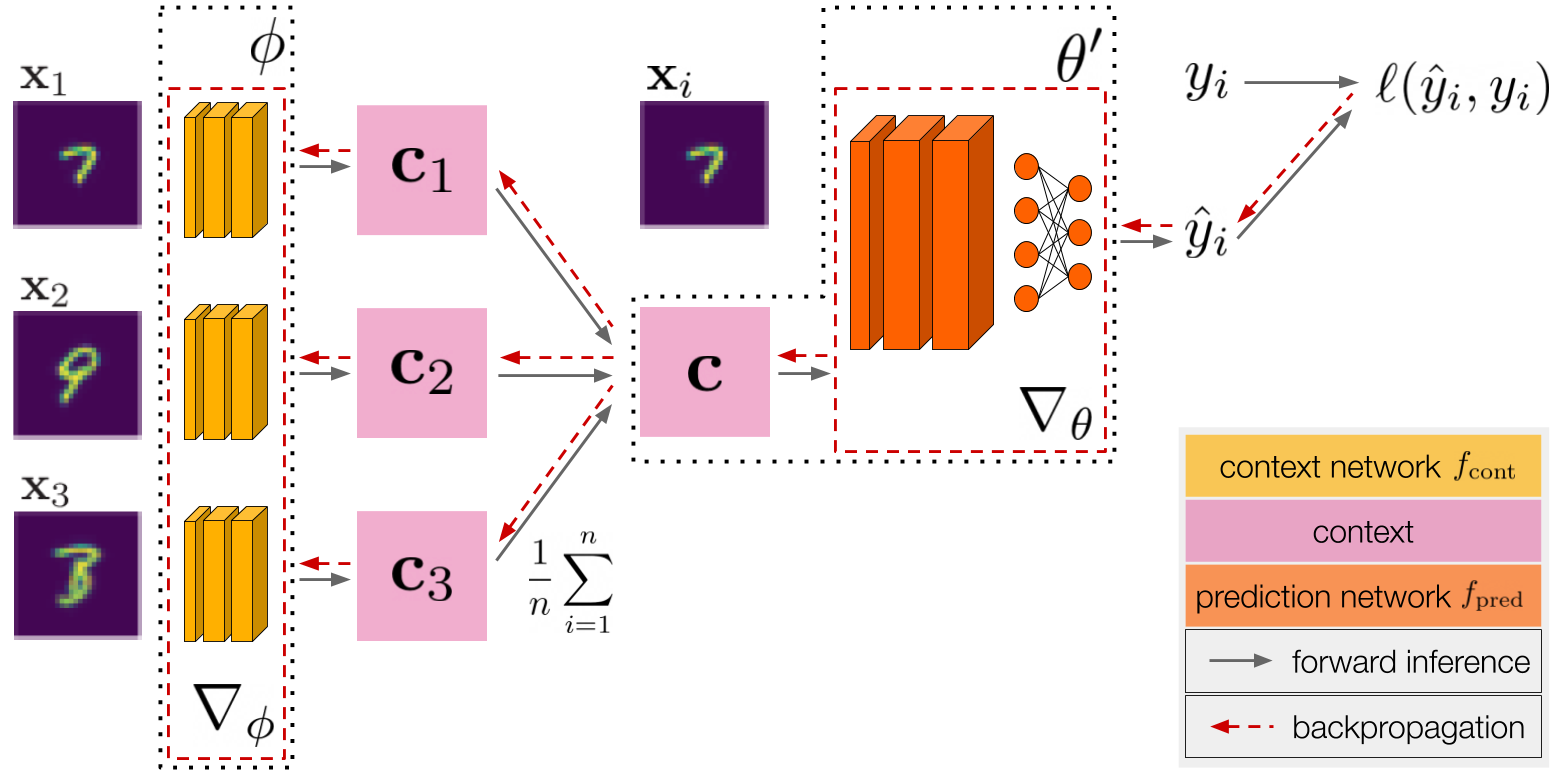}
    \caption{During inference for ARM-CML, the context network produces a vector $\context_k$ for each input image $\x_k$ in the batch, and the average of these vectors is used as the context $\context$ is input to the prediction network. This context may adapt the model by providing helpful information about the underlying test distribution, and this adaptation can aid prediction for difficult or ambiguous examples. During training, we compute the loss of the post adaptation predictions and backpropagate to update $\params$ and $\learnerparams$.}
    \label{fig:cml}
\end{wrapfigure}

A schematic of the ARM-CML method is presented in \autoref{fig:cml}. The post adaptation model parameters $\params'$ are $[\params,\context]$. Since we only ever use the model after adaptation, both during training and at test time, we can simply specify \mbox{$\model(\x;\params')=\prednet(\x,\context;\params)$}. Though not strictly necessary, we could define the behavior of $\model$ before adaptation, i.e., with unadapted parameters $\params$, as using a running average of the context computed throughout the course of training, similar to BN. We then also see that $\learner$ is a function that takes in $(\params,\x_1,\ldots,\x_K)$ and produces $\left[\params,\frac{1}{K}\sum_{k=1}^K\contextnet(\x_k;\learnerparams)\right]$. In the streaming setting, we keep track of the average context over the previous test points $\context$ and we maintain a counter $t$ of the number of test points seen so far.\footnote{An alternative to maintaining a counter $t$ is to use an exponential moving average, though we do not experiment with this.} When we observe a new point $\x$, we increment the counter and update the average context as $\frac{t}{t+1}\context+\frac{1}{t+1}\contextnet(\x;\learnerparams)$, and then we make a prediction on $\x$ using this updated context. Notice that, with this procedure, we do not need to store any test points after they are observed, and this procedure results in an equivalent context to ARM-CML in the batch test setting after observing $K$ data points.

In the streaming setting, ARM-BN is similar to ARM-CML, though slightly more complex due to the requirement of computing second moments. Denote the context after seeing $t$ test points as $\context=[\bm{\mu},\bm{\sigma}^2]$, the mean and variance of the BN layer activations on the points so far. Upon seeing a new test point, let $\mathbf{a}$ denote the BN layer activations computed from this new point, with size $h$. We then update $\context$ to be
\[
\left[\frac{ht}{h(t+1)}\bm{\mu}+\frac{\sum\mathbf{a}}{h(t+1)},\frac{ht}{h(t+1)}(\bm{\sigma}^2+\bm{\mu}^2)+\frac{\sum\mathbf{a}^2}{h(t+1)}-\left(\frac{ht}{h(t+1)}\bm{\mu}+\frac{\sum\mathbf{a}}{h(t+1)}\right)^2\right]\,.
\]
Again note that we do not store any test points and that we arrive at the same context as the batch test setting after observing $K$ data points.

For ARM-LL, in our experiments, we used $\alpha=0.1$ and 1 gradient step for both meta-training and meta-testing. Though we did not evaluate ARM-LL in the streaming setting, in principle this method can be extended to this setting by performing a gradient step with a smaller $\alpha$ after observing each test point. In an online fashion, similar to \citet{sun20icml}, we can continually update the model parameters over the course of testing rather than initializing from the meta-learned parameters for each test point.

\section{Contrasting with Prior Benchmarks}
\label{sec:supp-bench}

One aim of this work is to identify problems for which unlabeled adaptation is feasible, helpful, and potentially crucial, inspired by important real world problem settings such as federated learning. Thus, the problems we focus on will naturally differ from prior work in domain shift, which have different implicit and explicit goals when designing and choosing benchmarks. In particular, as discussed in \autoref{sec:intro}, many prior benchmarks assume the existence of a consistent input-output relationship across domains, for which various methods can be designed to try and better uncover this relationship. Compared to problems where adaptation is important, we can roughly characterize these benchmarks as having a conditional distribution $p(y\vert\x)$ that is more stable across domains and thus does not depend as much on the marginal $p(\x)$. As \citet{blanchard21jmlr} note informally, and as mentioned in \autoref{sec:prelim}, the less information the marginal provides about the conditional, the less we expect domain generalization strategies to improve over ERM. Indeed, \citet{gulrajani21iclr} provide a comprehensive survey of domain generalization benchmarks and find that, though ERM is sometimes outperformed by certain methods on certain benchmarks, ERM is competitive with the state of the art on average across the benchmarks.

\citet{gulrajani21iclr} also evaluated ARM-CML across the whole suite and found middling performance across most of the testbeds in the benchmark. This negative result provides further evidence that adaptation may not be well suited to these problems, at least in their standard formulations. Similarly, adaptation and ARM methods also do not improve performance on some of the \textsc{Wilds} domain generalization problems~\citep{koh21icml}, potentially due to the marginal $p(\x)$ not providing much information about $p(y\vert\x)$, or other factors such as the lack of training domains or shared structure between domains. One potentially interesting result that \citet{gulrajani21iclr} found was that ARM-CML did outperform ERM and all prior methods on one toy problem: the colored MNIST benchmark~\citep{arjovsky19}. For a non adaptive model, the goal as originally proposed in \citet{arjovsky19} is to disregard color and learn the invariant relationship between digits and labels. Irrespective of the original motivations, though, an adaptive model is in theory capable of learning a more flexible classification strategy for this problem, in that it may leverage information about the current domain in order to produce better predictions. Viewed this way, it becomes clear why ARM-CML can learn a more performant solution for the colored MNIST problem. This result on a toy problem provides further motivation for identifying and studying real world problems for which adaptation can be beneficial, alongside other benchmarks geared toward discovering invariances.

Specifically when viewing learning adaptation as a meta-learning problem, as in this work, we may pose additional hypotheses about a problem's desired properties. For example, in meta-learning, each task is viewed as a ``higher level'' data point, and this generally motivates constructing many different tasks so as to prevent the learner from overfitting to the tasks. We extend this intuition to our work in that our problems have tens to hundreds of domains, whereas the benchmarks in DomainBed have between 3 to 6 domains. Note that the overall dataset sizes are still comparable, so previous benchmarks typically also have orders of magnitude more data per domain. Depending on the scenario, it may be difficult to either collect data from many domains, or conversely it may be difficult to collect many data points from any single domain. For example, the FEMNIST dataset naturally contains hundreds of users each contributing at most hundreds of examples, but it would be difficult to collect orders of magnitude more data from any given user. These practical considerations should also factor into the choice of algorithm for solving any particular problem.

Prior testbeds used in group distributionally robust optimization~(DRO) typically also contain a small number of groups~\citep{sagawa20iclr}, and these testbeds also have a couple of other important differences. First, as discussed in \autoref{sec:experiments}, group DRO testbeds typically use the same training and test groups and measure worst case performance, which differs from domain generalization, meta-learning, and most problems considered in this work, which construct or hold out disjoint sets of domains for testing. Second, prior group DRO testbeds use label information to construct groups, in that data within each group will all have the same label. This is not an issue for non adaptive models, however, classification in this setup becomes much easier for adaptive models and particularly if training with an ARM method, as the model simply needs to learn to adapt to output a constant label. Thus, in this work, we identify and set up problems which are distinct from both prior work in domain generalization and group DRO, in order to properly evaluate ARM and prior methods in settings for which adaptation is beneficial.

\section{Additional Experimental Details}
\label{sec:supp-setup}

Code for \autoref{tab:results} results is available from \url{https://github.com/henrikmarklund/arm}.

In our experiments, we use several different computing clusters with either NVIDIA Titan X Pascal, RTX 2080 Ti, or V100 GPUs, and all experiments use 1 GPU. When reporting our results, we run each method across three seeds and reported the mean and standard error across seeds. Standard error is calculated as the sample standard deviation divided by $\sqrt{3}$. We checkpoint models after every epoch of training, and at test time, we evaluate the checkpoint with the best worst case validation accuracy. Training hyperparameters and details for how we evaluate validation and test accuracy are provided for each experimental domain below. All hyperparameter settings were selected in preliminary experiments using validation accuracy only.

We also provide details for how we constructed the splits for each dataset. These splits were designed without any consideration for the train, validation, and test accuracies of any method. All of these design choices were made either intuitively -- such as maintaining the original data splits for MNIST -- or randomly -- such as which users were selected for which splits in FEMNIST -- or with a benign alternate purpose -- such as choosing disjoint sets of corruptions with mostly different types.

\subsection{Rotated MNIST details}

We construct a training set of 32292 data points using 90\% of the original training set -- separating out a validation set -- by sampling and applying random rotations to each image. The rotations are not dependent on the image or label, but certain rotations are sampled much less frequently. Rotations of 0 through 20 degrees, inclusive, have 7560 data points each, 30 through 50 degrees have 2160 points each, 60 through 80 have 648, 90 through 110 have 324 each, and 120 to 130 have 108 points each.

We train all models for 200 epochs with mini batch sizes of $300$. We use Adam updates with learning rate $0.0001$. We construct an additional level of mini batching for our method as described in \autoref{sec:arm}, such that the batch dimensions of the data mini batches are $6\times50$ rather than just $300$, and each of the inner mini batches contain examples from the same rotation. We refer to the outer batch dimension as the \emph{meta batch size} and the inner dimension as the batch size. All methods are still trained for the same number of epochs and see the same amount of data. DRNN uses an additional learning rate hyperparameter for their robust loss, which we set to $0.01$ across all experiments~\citep{sagawa20iclr}.

We compute validation accuracy every 10 epochs. We estimate validation accuracy on each rotation by randomly sampling 300 of the held out 6000 original training points and applying the specific rotation, resampling for each validation evaluation. This is effectively the same procedure as the test evaluation, which randomly samples 3000 of the 10000 test points and applies a specific rotation.

We retain the original $28\times28\times1$ image dimensionality, and we divide inputs by 256. We use convolutional neural networks for all methods with varying depths to account for parameter fairness. For all methods that do not use a context network, the network has four convolution layers with 128 $5\times5$ filters, followed by $4\times4$ average pooling, one fully connected layer of size 200, and a linear output layer. Rectified linear unit (ReLU) nonlinearities are used throughout, and BN~\citep{ioffe15icml} is used for the convolution layers. The first two convolution layers use padding, and the last two convolution layers use $2\times2$ max pooling. For ARM-CML and the context ablation, we remove the first two convolution layers for the prediction network, but we incorporate a context network. The context network uses two convolution layers with 64 filters of size $5\times5$, with ReLU nonlinearities, BN, and padding, followed by a final convolution layer with 12 $5\times5$ filters with padding.

\subsection{FEMNIST details}

FEMNIST, and EMNIST in general, is a significantly more challenging dataset compared to MNIST due to its larger label space (62 compared to 10 classes), label imbalance (almost half of the data points are digits), and inherent ambiguities (e.g., lowercase versus uppercase ``o'')~\citep{cohen17}. In processing the dataset,\footnote{\url{https://github.com/TalwalkarLab/leaf/tree/master/data/femnist}.} we filter out users with fewer than 100 examples, leaving 262, 50, and 35 unique users and a total of 62732, 8484, and 8439 data points in the training, validation, and test splits, respectively. The smallest users contain 104, 119, and 140 data points, respectively. We keep all hyperparameters the same as for rotated MNIST, except we set the meta batch size for ARM methods to be 2, and we use stochastic gradient updates with learning rate 0.0001, momentum 0.9, and weight decay 0.0001. For DANN, we use Adam updates with learning rate $0.0001$ as stochastic gradient updates were unsuccessful for this method.

We compute validation accuracy every epoch by iterating through the data of each validation user once, and this procedure is the same as test evaluation. Note that all methods will sometimes receive small batch sizes as each user's data size may not be a multiple of 50. Though this may affect ARM methods, we demonstrate in \autoref{sec:experiments} that ARM-CML and ARM-BN can adapt using small batch sizes, such as in the streaming test setting. The network architectures are the same as the architectures used for rotated MNIST, except that, when applicable, the last layer of the context network has only 1 filter of size $5\times5$.

\subsection{CIFAR-10-C and Tiny ImageNet-C details}

For both CIFAR-10-C and Tiny ImageNet-C, we construct training, validation, and test sets with 56, 17, and 22 domains, respectively. Each domain is based on type and severity of corruption. We split domains such that corruptions in the training, validation, and test sets are disjoint. Specifically, the training set consists of Gaussian noise, shot noise, defocus blur, glass blur, zoom blur, snow, frost, brightness, contrast, and pixelate corruptions of all severity levels. Similarly, the validation set consists of speckle noise, Gaussian blur, and saturate corruptions, and the test set consists of impulse noise, motion blur, fog, and elastic transform corruptions of all severity levels. For two corruptions, spatter and JPEG compression, we include lower severities (1-3) in the training set and higher severities (4-5) in the validation and test sets. In this way, we are constructing a more challenging test setting, in which the test domains are not sampled identically as the training domains, since the corruption types are largely different between the two sets. For the training and validation sets, each domain consists of 1000 images for CIFAR-10-C and 2000 images for Tiny ImageNet-C, giving training sets of size 56000 and 112000, respectively. We use the full test set of 10000 images for each domain, giving a total of 220000 test images for both datasets.

In these experiments, we use a support size of 100 and meta batch size of 3. For CIFAR-10-C, we use the same convolutional network architecture as for rotated MNIST and FEMNIST, except for the first layer which needs to be modified to handle RGB images. For Tiny ImageNet-C, we fine tune ResNet-50~\citep{he16cvpr} models pretrained on ImageNet. The context ablation and ARM-CML additionally use small convolutional context networks, and the learned loss ablation and ARM-LL use small fully connected loss networks. For this domain, we further incorporate BN adaptation into the context ablation and ARM-CML, as we found this technique to generally be very helpful when dealing with image corruptions. The images are first normalized by the ImageNet mean and standard deviation. For CIFAR-10-C, we train models from scratch for 100 epochs, and for Tiny ImageNet-C we fine tune for 50 epochs. We use stochastic gradient updates with learning rate 0.01, momentum 0.9, and weight decay 0.0001. We evaluate validation accuracy after every epoch and perform model selection based on the worst case accuracy over domains. We perform test evaluation by randomly sampling 3000 images from each domain and computing worst case and average accuracy across domains.

\section{Additional Experiments}
\label{sec:supp-exp}

\subsection{Additional comparisons}

\begin{table}[t]
    \centering
    \caption{Comparing to DANN~\citep{ganin15icml} as an unsupervised domain adaptation~(UDA) method, in which the particular test domain is known at training time. Note that this involves retraining models for each test evaluation, and ARM-CML is still more performant by leveraging meta-training and adaptation.}
    \label{tab:uda}
    \begin{tabular}{lcc}
        \toprule
                        & \tc{\textbf{Rotated MNIST}} \\
                          \cmidrule(lr){2-3}
        \textbf{Method} & \textbf{WC}      & \textbf{Avg} \\
        \midrule
        DANN~(DG)       & $78.8\pm0.8$     & $94.9\pm0.1$ \\
        DANN~(UDA)      & $82.4\pm1.6$     & $94.9\pm0.2$ \\
        ARM-CML         & $\*{88.0\pm0.8}$ & $\*{96.3\pm0.4}$ \\
        \bottomrule
    \end{tabular}
    \vspace{0.25em}
\end{table}

\begin{table}[t]
    \centering
    \caption{Comparing to a modified version of ARM-CML with probabilistic contexts, similar to~\citet{kumagai18}. The standard formulation of ARM-CML performs better on rotated MNIST and FEMNIST, possibly due to the objective purely encouraging predictive accuracy.}
    \label{tab:prob}
    \begin{tabular}{lcccc}
        \toprule
                                    & \tc{\textbf{Rotated MNIST}}         & \tc{\textbf{FEMNIST}} \\
                                      \cmidrule(lr){2-3}                    \cmidrule(lr){4-5}
        \textbf{Method}             & \textbf{WC}      & \textbf{Avg}     & \textbf{WC}      & \textbf{Avg} \\
        \midrule
        ARM-CML                     & $\*{88.0\pm0.8}$ & $\*{96.3\pm0.4}$ & $\*{70.9\pm1.4}$ & $\*{86.4\pm0.3}$ \\
        ARM-CML w/ prob. $\context$ & $82.6\pm0.6$     & $93.8\pm0.5$     & $65.1\pm2.5$     & $84.7\pm0.7$ \\
        \bottomrule
    \end{tabular}
    \vspace{0.25em}
\end{table}

In \autoref{tab:uda} and \autoref{tab:prob}, we provide additional comparisons to unsupervised domain adaptation~(UDA) methods and zero shot domain adaptation methods, respectively. A number of methods have been proposed for UDA, and for simplicity, we compare to DANN~\citep{ganin15icml}, which we evaluated in \autoref{sec:experiments} as a domain generalization algorithm but was originally proposed for UDA. When faced with multiple test shifts, UDA methods run training separately for each shift, as they assume access to unlabeled samples from the test distribution at training time. For rotated MNIST, where there are 14 test groups, evaluating DANN as a UDA method involved 42 separate training runs, as we still used 3 training seeds per test evaluation. We see \autoref{tab:uda} that DANN in this setting performs better in terms of worst case accuracy, which is not surprising given each model's ability to specialize to a particular test domain. However, by leveraging meta-training and adaptation, ARM-CML still performs the best on this problem.

As noted in \autoref{sec:related}, \citet{kumagai18} propose a method for zero shot domain adaptation that is quite similar to ARM-CML, with the primary high level difference being that, in their method, the contexts are treated as probabilistic latent variables. We thus evaluate a variant of ARM-CML in which we placed a unit Gaussian prior independently on each dimension of the context $\context$ and optimized an evidence lower bound. In \autoref{tab:prob}, we see that this variant generally performed worse than the original formulation of ARM-CML, possibly due to the objective balancing between satisfying a restrictive prior and optimizing for predictive accuracy.

\subsection{Additional results with loosened assumptions}

\begin{figure}
    \centering
    \includegraphics[width=0.4\textwidth]{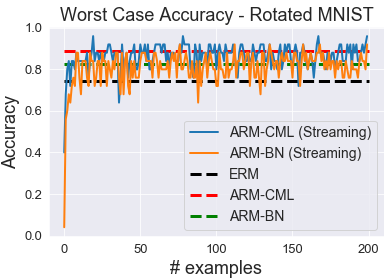}
    \includegraphics[width=0.4\textwidth]{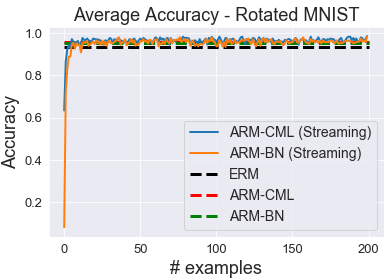}
    \caption{On rotated MNIST, ARM methods reach strong performance in the streaming setting after fewer than 10 data points, again despite meta-training with batch sizes of 50.}
    \label{fig:more-streaming}
\end{figure}

In \autoref{fig:more-streaming}, we include results for the rotated MNIST problem in the test streaming setting. We can see the same general trend as for Tiny ImageNet-C, where the models trained via ARM methods are able to adapt successfully, and in this easier domain these models require fewer than 10 test inputs to reach their performances reported in \autoref{tab:results}, where adaptation is performed with batches of 50 points.

In the case of unknown domains, one option is to use unsupervised learning techniques to discover domain structure in the training data. To test this option, we focus on rotated MNIST and ARM-CML, which performs the best on this dataset, and train a variational autoencoder~(VAE)~\citep{kingma14iclr,rezende14icml} with discrete latent variables~\citep{jang17iclr,maddison17iclr} using the training images and labels.
\begin{wrapfigure}{r}{0.53\linewidth}
    \centering
    \includegraphics[width=0.9\linewidth]{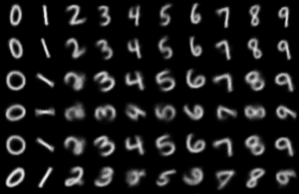}
    \caption{VAE samples conditioned on different values of $y$ (x axis) and $c$ (y axis). The VAE learns to use $c$ to represent rotations.}
    \label{fig:learned}
\end{wrapfigure}
We define the latent variable, which we denote as $c$ to differentiate from the domain $z$, to be Categorical with 12 possible discrete values, which we purposefully choose to be smaller than the number of rotations. The VAE is not given any information about the ground truth $z$; however, we weakly encode the notion that $c$ is independent of $y$ by conditioning the decoder on the label. We use the VAE inference network to assign domains to the training data, and we run ARM-CML using these learned domains. In \autoref{tab:learned}, we see that ARM-CML in this setting outperforms ERM and is competitive with TTT, which as discussed earlier encodes a strong inductive bias, via rotation prediction, for solving this task. \autoref{fig:learned} visualizes samples from the VAE for different values of $y$ and $c$, which shows that the VAE learns to encode rotation information using $c$. This result suggests that, when domain information is not provided, a viable approach may be to learn domains which then enables the use of ARM methods.

\begin{table}
    \centering
    \small
    \caption{Using learned domains, ARM-CML outperforms ERM and matches the performance of TTT on rotated MNIST. This result may be improved by techniques for learning more diverse domains.}
    \label{tab:learned}
    \begin{tabular}{lcc}
    \toprule
    \textbf{Method} & \textbf{WC}      & \textbf{Avg} \\
    \midrule
    ERM             & $74.5\pm1.4$     & $93.6\pm0.4$ \\
    TTT             & $\*{81.1\pm0.3}$ & $\*{95.4\pm0.1}$ \\
    ARM-CML         & $\*{81.7\pm0.3}$ & $\*{95.2\pm0.3}$ \\
    \bottomrule
    \end{tabular}
    \vspace{0.25em}
\end{table}

\end{document}

%% file: defs.tex
\newcommand{\context}{\mathbf{c}}
\newcommand{\contextnet}{\text{f}_{\text{cont}}}

\newcommand{\E}{\mathbb{E}}
\newcommand{\fullmodel}{f}
\newcommand{\inspace}{\mathcal{X}}
\newcommand{\learner}{h}
\newcommand{\learnerparams}{\phi}
\newcommand{\lossnet}{\text{f}_{\text{loss}}}
\newcommand{\model}{g}

\newcommand{\outspace}{\mathcal{Y}}
\renewcommand{\P}{\mathcal{P}}
\newcommand{\params}{\theta}
\newcommand{\paramsspace}{\Theta}
\newcommand{\prednet}{\text{f}_{\text{pred}}}

\newcommand{\x}{\mathbf{x}}
\newcommand{\xy}{\mathbf{x}y}